%% file: main.tex
\documentclass[11pt]{article}

\usepackage[margin=1in]{geometry}
\usepackage{amsmath,amssymb,amsthm,mathtools}
\usepackage{graphicx}
\usepackage{booktabs}
\usepackage{microtype}
\usepackage{anyfontsize} 
\usepackage{enumitem}
\usepackage{url}
\usepackage{xcolor}

\usepackage[authoryear,round]{natbib}
\setcitestyle{citesep={,}}

\usepackage[hidelinks]{hyperref}

\usepackage[nameinlink,capitalise]{cleveref}

\usepackage{authblk}

\newcommand{\pmtf}{\mathrm{PM2.5}}
\newcommand{\excessexposure}{\mathrm{DEE}}

\DeclareMathOperator*{\argmin}{arg\,min}

\newcounter{suppsection}
\renewcommand{\thesuppsection}{S\arabic{suppsection}}
\newcommand{\suppsection}[1]{%
  \refstepcounter{suppsection}%
  \section*{Supplementary Section \thesuppsection: #1}%
  \addcontentsline{toc}{section}{Supplementary Section \thesuppsection: #1}%
}

\crefname{suppsection}{Supplementary Section}{Supplementary Sections}
\Crefname{suppsection}{Supplementary Section}{Supplementary Sections}

\newcommand{\datastatement}{\section*{Data Availability Statement}}
\let\origappendix\appendix
\renewcommand{\appendix}[1][]{\origappendix}

\title{Are Hourly PM2.5 Forecasts Sufficiently Accurate to Plan Your Day?\\
Individual Decision Making in the Face of Increasing Wildfire Smoke}

\author[1]{Renato Berlinghieri\thanks{ denotes equal contribution. Correspondence: \texttt{renb@mit.edu, dburt@mit.edu.}}}
\author[1]{David R.~Burt\protect\footnotemark[1]}
\author[2]{Paolo Giani}
\author[2]{Arlene Fiore}
\author[1]{Tamara Broderick}

\affil[1]{MIT LIDS and MIT-IBM Watson AI Lab}
\affil[2]{MIT EAPS}

\date{} 

\begin{document}
\maketitle

\begin{abstract}
Wildfire frequency is increasing as the climate changes, and the resulting air pollution poses health risks. Just as people routinely use hourly weather forecasts to plan their day’s activities around precipitation, reliable hourly air quality forecasts could help individuals reduce their exposure to air pollution. In the present work, we evaluate six existing forecasts of ground-level fine particulate matter (PM2.5) within the continental United States during the 2023 fire season. We include forecasts using physical simulation, ensembling, and artificial intelligence. 
We focus our evaluation on individual decisions, such as (1) whether to go outside on a day with potentially high PM2.5 or (2) when to go outside for the lowest PM2.5 exposure.
Our evaluation consists of both visualizations of hourly PM2.5 forecasts in particular locations as well as metrics summarizing forecast skill for the two tasks above. 
As part of our analysis, we introduce a new evaluation metric for the task of deciding when to go outside. We find meaningful room for improvement in PM2.5 forecasting, which might be realized by improving physical models, incorporating more data sources, and using artificial intelligence tools.
\end{abstract}

\section{Introduction}\label{sec:introduction}
\input{sections/introduction}

\section{Prior PM2.5 Forecast Evaluations}\label{sec:related_work}
\input{sections/related_work}

\section{Data Description}\label{sec:study_scope}
\input{sections/study_scope}

\section{Methods for Evaluating Forecasts for Individual Decision Making}\label{sec:metrics}
\input{sections/metrics}

\section{Evaluating Existing Forecasts for Individual Decision Making}\label{sec:forecast_evaluation}
\input{sections/forecast_evaluation}

\section{Ensembling and Artificial Intelligence Can Improve Forecasts, But Not Automatically}\label{sec:discussion} 

Our analysis of PM2.5 forecasts during the 2023 wildfire season in CONUS demonstrates that current models provide useful but insufficiently reliable guidance for individuals seeking to minimize health risks in the face of wildfires and other smoke events. We find that, when used to plan \emph{when} to go outside during a high-pollution day, many forecasts (especially GEOS-CF, CAMS, and NAQFC) can substantially cut excess exposure to PM2.5. However, for deciding \emph{whether} to go outside today, our analysis suggests that using the PM2.5 value in the morning would guide an individual about as well as, or better than, any forecast we analyzed here. We believe there remains meaningful room for improvement, and we hope that the present paper can help establish a framework for future evaluations as forecasts continue to improve. One promising avenue comprises improved data inputs, improved meteorology, or improved chemistry within traditional forecasting approaches. \citet{DevelopmentofthenextgenerationairqualitypredictionsystemintheUnifiedForecastSystemframeworkEnhancingpredictabilityofwildfireairqualityimpacts} suggested that these factors were crucial to the large improvement in NAQFC forecasting that they report between AQMv6 (the version of NAQFC that was operational in 2023 and that we include here) and AQMv7 (the currently operational version of NAQFC since 2024). \citet{MultiagencyEnsembleForecastofWildfireAirQualityintheUnitedStatesTowardCommunityConsensusofEarlyWarning} suggest that ensembles of existing forecasts can also lead to improvements in forecast quality. Of the forecasts studied by \citet{MultiagencyEnsembleForecastofWildfireAirQualityintheUnitedStatesTowardCommunityConsensusofEarlyWarning}, the one they released is what we presently studied as HAQES; HAQES takes the (unweighted) mean of existing forecasts. Our analysis illustrates that this mean need not lead to a substantial improvement in forecast quality. \citet{MultiagencyEnsembleForecastofWildfireAirQualityintheUnitedStatesTowardCommunityConsensusofEarlyWarning} also considered an alternative approach that weights forecasts with weights that need not be equal or add to one, but such that the weight applied to each forecast is fixed going forward. Since no forecast consistently outperforms all others in our analysis, we expect even more progress to be possible with weighting approaches that allow the weights given to each forecast to depend on meteorological or other covariates. Under our latter proposal, each forecast could be given more weight in conditions in which it performs well, and less weight in situations in which it falters. However, the case of San Francisco suggests that even sophisticated ensembling strategies may fall short during certain extreme smoke events, where a persistence forecast surprisingly outperforms more complex models. AI and machine learning approaches represent another promising direction due to their flexibility in incorporating diverse data sources. However, our analysis demonstrates that even state-of-the-art AI approaches like Aurora can underperform compared to traditional physical forecasts in predicting air quality, particularly for PM2.5. Nonetheless, as we argue next, it is possible within an AI framework to surmount three key limitations of Aurora: (1) lack of direct use of wildfire information, (2) lack of training on ground-truth data, and (3) overly coarse temporal resolution of predictions. First, Aurora currently integrates wildfire information only through CAMS predictions, which Aurora takes as input. But there exist additional sources of data that should be informative about PM2.5 spread due to wildfires: e.g., remote-sensing information about the location and intensity of wildfires as well as biomass inventory data. Second, to create their foundation model with a focus on weather prediction, Microsoft initially trained Aurora on a large earth system analysis and reanalysis dataset. To adjust the foundation model for the purposes of PM2.5 prediction, Microsoft then did a second (or `fine-tuning'') round of training on CAMS analysis and reanalysis data. But \citet{LEE2024120833} showed that CAMS reanalysis data can have significant biases compared to AirNow observations, particularly when wildfires lead to elevated PM2.5 levels. We expect that fine-tuning on directly observed air quality data (such as data from monitors reporting to AirNow) could substantially improve predictions. Third, Aurora makes PM2.5 predictions only every 12 hours. But PM2.5 arising from wildfire smoke typically varies over spatial scales of miles and temporal scales of hours, which should facilitate effective modeling given adequate training data. 

Finally, extreme events such as the air pollution events seen in the eastern United States in June 2023 are, by definition, rare. And their frequency is expected to be non-stationary (namely, growing) in time. As such, we expect that traditional AI approaches that recognize patterns in assumed-stationary processes can be improved. In particular, we expect that accurate forecasts of extreme smoke events will require interdisciplinary collaboration, incorporating robust physical insights into atmospheric processes alongside AI tools. 

\section*{Acknowledgments}
This work was supported in part by a Social and Ethical Responsibilities of Computing (SERC) seed grant. Renato Berlinghieri, David R.\ Burt, and Tamara Broderick were supported by the MIT-IBM Watson AI Lab. Paolo Giani and Arlene M.\ Fiore were supported by the Climate Grand Challenge BC3 project, and Arlene M.\ Fiore was supported by NASA HAQAST 80NSSC21K0509. The authors thank Daniel Tong and Barry Baker for their help retrieving NAQFC data. The authors would also like to thank Wessel P.\ Bruinsma and Megan J.\ Stanley for their help accessing Microsoft Aurora data and J. Nathan Matias for helpful discussions.

\datastatement

\newcommand{\codeurl}{\url{https://github.com/DavidRBurt/BAMS-PM25-Forecasting/}}
\newcommand{\airnowlink}{\url{https://files.airnowtech.org/}}
\newcommand{\geoscflink}{\url{https://portal.nccs.nasa.gov/datashare/gmao/geos-cf/v1/forecast/}}
\newcommand{\haqeslink}{\url{https://disc.gsfc.nasa.gov/datasets/HAQES_NA_PM25_TOT_1/summary?keywords=HAQES\%20pm2.5}}
\newcommand{\hrrrlink}{\url{https://registry.opendata.aws/noaa-hrrr-pds/}}
\newcommand{\naqfclink}{\url{https://registry.opendata.aws/noaa-nws-naqfc-pds/}}
\newcommand{\camslink}{\url{https://ads.atmosphere.copernicus.eu/datasets/cams-global-atmospheric-composition-forecasts?tab=overview}}

Our code for downloading the forecasts and AirNow data is publicly available at \codeurl. The AirNow data is available at \airnowlink; the GEOS-CF forecasts are available at \geoscflink; the NAQFC forecasts are available at \naqfclink; the HRRR-Smoke forecast are available at \hrrrlink; and the HAQES forecast are available at \haqeslink. These datasets are all US government works. The CAMS forecast data is available at \camslink; the license for this data is \url{https://apps.ecmwf.int/datasets/licences/cams/}. The Aurora predictions are under a CDLA2.0 license. Microsoft generated Aurora forecasts using Copernicus Atmosphere Monitoring Service Information (2025).

\bibliographystyle{plainnat} 
\bibliography{references}

\appendix[A]
\section*{Appendix}
\input{sections/appendix/setup_details}

\clearpage
\renewcommand\thesection{\arabic{section}}
\renewcommand\thesubsection{\arabic{section}.\arabic{subsection}}
\setcounter{section}{0}
\setcounter{subsection}{0}

\suppsection{Air Quality Forecasts We Consider}\label{sec:forecasts_supplement}
\input{sections/supplement/forecasts}

\end{document}

%% file: sections/introduction.tex
Wildfire smoke is an increasingly pressing issue, with climate change driving more frequent and intense fires \citep{burke_contribution_2023,Liu2016particulate,abatzoglou2016impact}. Wildfires release fine particulate matter (PM2.5), which can travel great distances and degrade air quality. In the United States, fire smoke often drives extreme levels of fine particulate matter (for example, exceeding the EPA 24-hour exposure guidance of 35 $\mu g/m^3$) \citep{Liu2016particulate}. Researchers have linked exposure to elevated PM2.5 levels to adverse health outcomes, including reduced birth weight \citep{holstius_birth_2012} and cardiopulmonary diseases \citep{adetona_review_2016, deflorio-barker_cardiopulmonary_2019,Dominici2006Fine}. Given these health risks, people want to reduce their exposure to outdoor PM2.5, and air pollution forecasts offer a potential mechanism for modifying individual behavior to achieve this reduction.

Air pollution forecasts, analogous to weather forecasts, offer predictions about air quality conditions in advance. Individuals can use these forecasts to plan their activities, such as deciding whether to exercise outdoors, carry a face mask, keep windows closed, or use air filtration systems. However, for these forecasts to be useful in practice, they must be reliable at the level of individual decision making. For instance, an individual deciding whether to go outside at all today would like to reliably know whether PM2.5 levels will be high today. An individual choosing the best time to go outside would like to reliably know when PM2.5 levels will be lowest.


In this paper, we evaluate six existing PM2.5 forecasts in the context of individual decision making: four based on meteorological and chemical simulation, one ensemble of forecasts, and one artificial intelligence (AI) forecast. 
Our evaluation spans the contiguous United States (CONUS) in the 2023 fire season. We advocate for using a mix of metrics and visualizations when conducting this evaluation and propose a new metric, mean excess exposure (MEE). MEE quantifies the additional PM2.5 exposure an individual would face when relying on a particular forecast compared to the true PM2.5 levels (which are not known in advance). We find that existing forecasts provide some useful information to individuals about timing of PM2.5 levels; however, no single forecast consistently and accurately predicts which days will have elevated PM2.5 or which hours will have lowest PM2.5 within a day. 
We conclude that substantial room for improvement remains, which is likely achievable with continued efforts to refine physical models, integrate additional data sources, and design appropriate AI methods. 


%% file: sections/related_work.tex
Evaluations of PM2.5 forecasts to date primarily rely on aggregate metrics such as mean absolute error, mean absolute bias, or a correlation coefficient \citep[e.g.][]{campbell2022development,zhang2022development}. 
While these metrics are useful in other contexts, averaging errors over long periods may obscure performance during high-pollution events, which are critical for individual decision making. 
Conversely, \citet{chow_high-resolution_2022,ye2021evaluation} each consider a single wildfire event.
But one might be concerned about how well conclusions drawn from any single event would generalize to broader forecasting performance.


Both \citet{yao_evaluation_2013,ainslie_operational_2022} have evaluated PM2.5 forecasts for public health decision making. First, \citet{yao_evaluation_2013} examined correlations between predicted smoke exposure levels and asthma-related physician visits during wildfire events. This approach can help public health officials decide resource allocation during the fire season. By contrast, we are interested in cases where individuals, who need not be public health officials or suffering an acute medical event, are planning daily activities in light of ambient PM2.5.
Second, \citet{ainslie_operational_2022} tested the capability of FireWork \citep{chen2019firework}, an operational Canadian wildfire air quality model, to predict days with PM2.5 levels above specific thresholds during the 2016--2018 Canadian fire seasons. We take a similar approach when evaluating how well a forecast can help an individual plan whether to go outside today; however, we compare and evaluate six different forecasts, including recent AI and ensemble forecasts. Moreover, we provide a novel evaluation of forecast quality for planning within a single day.
Finally, we observe that not only air quality forecasts but also the frequency and severity of wildfire events have changed substantially since the fire seasons (2010, 2016--2018) analyzed by \citet{yao_evaluation_2013,ainslie_operational_2022}, so a new evaluation is needed.

%% file: sections/study_scope.tex
We next detail the range of forecasts, urban areas, and time period in our evaluation. 

\subsection{Air Quality Forecasts}
\label{sec:forecasts}
We evaluate six forecasts that account for wildfire smoke sources of PM2.5. When multiple versions of a forecast are available, we use the latest version active during the 2023 fire season. See \cref{sec:forecasts_supplement} for more detailed information about each forecast.
\begin{enumerate}
    \item High-Resolution Rapid Refresh (HRRR) is a numerical weather prediction model released and maintained by the National Oceanic and Atmospheric Administration (NOAA) \citep{dowell2022hrrrsmoke}. HRRR-Smoke is a module integrated into HRRR focused on modeling smoke emissions and transport.
    \item The Copernicus Atmosphere Monitoring Service (CAMS) is an air quality and atmospheric composition forecast system developed by the European Centre for Medium-Range Weather Forecasts (ECMWF) as part of the ECMWF Integrated Forecast System \citep{ecmwf2023cams}.
    \item The Global Earth Observing System Composition Forecasting (GEOS-CF) model is developed and maintained by NASA's Global Modeling and Assimilation Office (GMAO) \citep{keller2021geoscf}.
    \item The National Air Quality Forecast Capability (NAQFC), developed and maintained by the National Weather Service, integrates the North American Mesoscale (NAM) Forecast System with the Community Multiscale Air Quality (CMAQ) modeling system \citep{davidson2008naqfc,stajner2012naqfcpm}.
    \item The Hazardous Air Quality Ensemble System (HAQES) \citep{MultiagencyEnsembleForecastofWildfireAirQualityintheUnitedStatesTowardCommunityConsensusofEarlyWarning} combines predictions from HRRR-Smoke \citep{dowell2022hrrrsmoke}, GEFS-Aerosols \citep{zhang2022development}, NAQFC \citep{campbell2022development}, GEOS-FP \citep{gelaro2017modern} and NAAPS \citep{CHRISTENSEN19974169,gmd-9-1489-2016}.
    \item Aurora is a state-of-the-art AI foundation model for weather forecasting \citep{bodnar_foundation_2025}. Microsoft trained Aurora using a large amount of earth system data, and they fine-tuned a version of Aurora for air pollution forecasting at a 12-hour time resolution.  
\end{enumerate}
To help us understand the performance of these forecasts, we also evaluate the persistence baseline suggested by \citet{ainslie_operational_2022}; this baseline simply predicts that air quality will remain constant for the next 24 hours. We construct the starting points of this baseline from monitor data we accessed through the EPA \citep{airnow2024} --- specifically, those monitors that report to the EPA AirNow system.


\subsection{Urban Areas}
First, we choose the two urban areas with the largest population from each division of the U.S.\ Census Bureau (New England,
Middle Atlantic, East North Central, West North Central, South Atlantic, East South Central, West, South Central, Mountain, Pacific). Second, we include all urban areas among the 25 most populated urban areas nationwide in the 2020 Census. In total, we examine 30 urban areas, listed in \cref{app:metro-areas}. We calculate the AirNow PM2.5 value for each urban area as an average of AirNow monitor values near (within a 10km radius) the latitude and longitude coordinates of the urban area provided in the U.S.\ Census Gazetteer \citep{census}, with some exceptions to handle cases when many or no monitors are within a 10km radius; we provide details of how we handle these cases in \cref{app:spatial-averaging}.

\subsection{Time Period, Temporal Resolution, and Forecast Timing}

In order to include most of what would typically be considered the fire season in the continental United States and Canada, we study the period from May 1 to October 31, 2023 (inclusive). For each urban area, we exclude a date if either the AirNow readings or any forecast is missing. As a result, we exclude 489 of 5,520 possible urban area--days (8.9\%) and analyze the remaining 5,031 days. 120 excluded days represent missing Aurora forecasts during October 28--31. The remaining excluded days correspond to missing AirNow readings; Nashville has the most missing data (130 days), including all of May 1--August 26.

We use an hourly time resolution for both AirNow and, when available, forecasts. When hourly forecasts are not available, we treat the forecast as constant until the next prediction. For example, if a forecast with a 12-hour time resolution is initially made at midnight and predicts a PM2.5 concentration of 6 $\mu g/ m^3$ at noon and 12 $\mu g/ m^3$ at midnight the following day, we treat it as predicting 6 $\mu g/ m^3$ for hours from 1am to noon and 12 $\mu g/ m^3$ from 1pm to midnight.

In order to choose an initial time and forward time period when evaluating each forecast, we imagine an individual who, in the morning, wants to make plans for the next 24 hours. As a result, we generally use forecasts initialized at 12UTC, with predictions ranging from 13UTC on the same day to 12UTC the following day. The exact details vary slightly based on when particular forecasts run, and how far forward they predict; see \cref{app:forecast-timings}.

%% file: sections/metrics.tex
We evaluate and compare air quality forecasts with two critical decision-making scenarios in mind: (1) whether an individual should go outside at all during an extreme event, and (2) when to go out during the day to minimize exposure to harmful particulate matter.

\subsection{Visualizing Smoke Events}
Analogous to how individuals often use visualizations of hourly weather forecasts when making decisions, visualizations of PM2.5 forecasts reflect how an individual might want to use forecasts for making decisions in practice. Therefore, our first evaluation compares forecast visualizations to a visualization of an ideal (but unknown) forecast, namely one that uses the true PM2.5 values (which are not available at the time of forecasting).
Moreover, while numerical metrics offer concise summaries, they inherently simplify complex temporal and spatial details. Thus, we begin by advocating that evaluations include visual inspection of forecast data during extreme events to capture qualitative nuances.
Presently, we select significant smoke events from the 2023 fire season in three cities. Among other benefits, plotting highlights whether forecasts capture the magnitude of extreme pollution, the timing of pollution peaks and troughs, and the presence of spurious peaks. 
However, given the impracticality of visualizing every urban area and event, we also provide quantitative summaries below.

\subsection{Should I go out today? Identification of High-pollution Days}
\label{sec:high_pollution_setup}
After consulting the morning forecast, an individual in a sensitive group might decide to work from home if exposure during any hour of the day is expected to exceed a certain threshold, such as the Environmental Protection Agency's 24-hour limit of 35 $\mu g/m^3$. 
We summarize how well-founded these decisions would have been with a confusion matrix: a table showing true positives (days forecast to have high pollution that truly had high pollution), false positives, false negatives, and true negatives. We also report precision (what fraction of predicted high-pollution days truly had high pollution) and recall (what proportion of days with truly high pollution were predicted by the forecast).

For a forecast to be useful for identifying high-pollution days, it should have a substantially higher recall than the persistence baseline: that is, it should identify more days that will have elevated PM2.5 than a person can identify by checking the current PM2.5. Additionally, the forecast should not have a precision that is so low that a person would lose confidence in it; e.g., if most of its high-pollution predictions are false, an individual might cease to trust it.

\begin{figure}
    \centering
    \includegraphics[width=\linewidth]{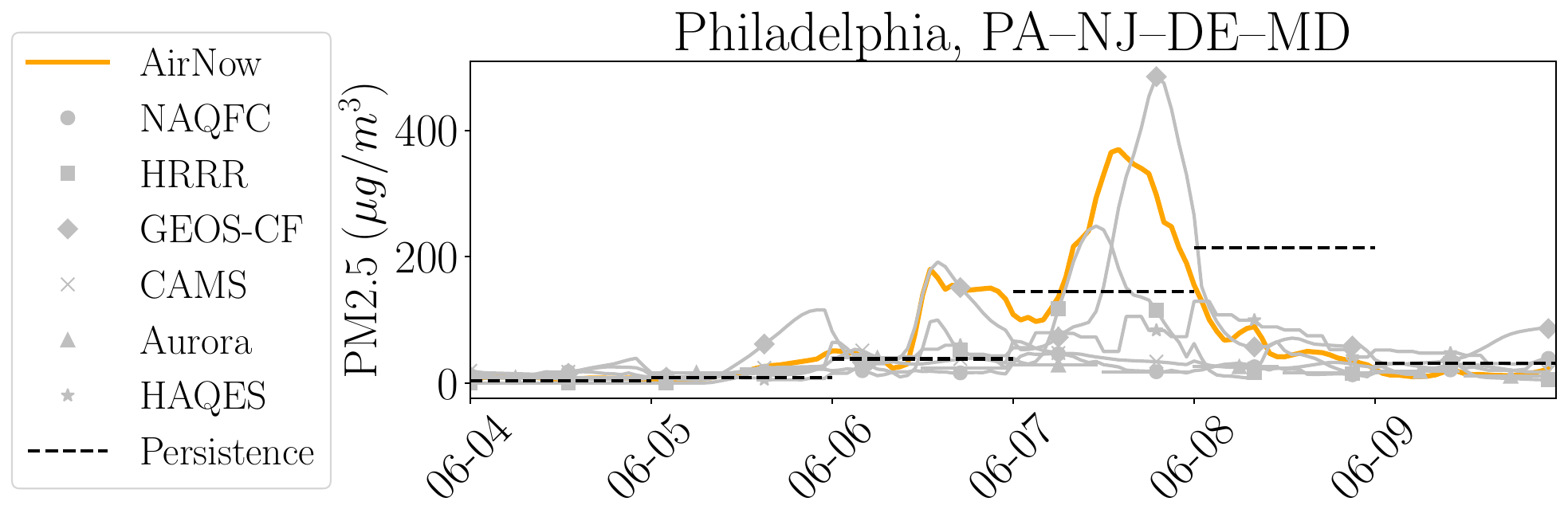}
    \caption{Air quality measurements obtained from AirNow (orange), a persistence baseline (black, dashed line) and 6 forecasts (gray, varying markers) in Philadelphia during June 4--9, 2023.}
    \label{fig:Philadelphia-June05-09}
\end{figure}


\subsection{When should I go out during the day? Mean Excess Exposure} We imagine someone who needs to spend an hour outside, is flexible about when to go outside, wants to minimize PM2.5 exposure, and will use the morning forecast to make their decision. To evaluate forecasts in this scenario, we define the \emph{mean excess exposure} (MEE); MME quantifies the additional PM2.5 exposure incurred by relying on a forecast compared to perfect knowledge. 
More precisely, to compute MEE, we first let $\pmtf_{d,s}(h)$ represent the true PM2.5 concentration at location $s$, day $d$, and hour $h$. Similarly, let $\widehat{\pmtf}_{d,s}(h)$ be the prediction of a particular forecast. Then we can define the \emph{day's excess exposure} ($\excessexposure$) for a particular forecast, day $d$, and location $s$ as the difference between PM2.5 exposure at the best hour indicated by the forecast and the true minimum hourly PM2.5 exposure:
\begin{align}
    \smash{\excessexposure(d, s) = \pmtf_{d,s}(\argmin_{h \in \{1, \dots, 24\}} \widehat{\pmtf}_{d,s}(h)) - \min_{h \in \{1, \dots, 24\}} \pmtf_{d,s}(h).}
\end{align}
Finally, the mean excess exposure of a forecast is the average of the DEE across a specified collection of days $\cal{T}$: 
$\frac{1}{|\cal{T}|}\sum_{d \in \cal{T}} \excessexposure(d, s)$. Since the forecast appears only within an argmin, note that shifting or scaling forecast values uniformly over time has no effect on its MEE. When a forecast predicts several hours to have the same exposure, or for forecasts with coarser time resolution (HAQES, Aurora, persistence), we define the MEE to be the expected MEE of a uniformly random hour selected from the set of hours that the forecast predicts to have lowest PM2.5.

%% file: sections/forecast_evaluation.tex
In the analyses below, we find that forecasts contain some useful information about air pollution, particularly in terms of mean excess exposure. For many smoke events, at least one forecast predicts elevated PM2.5 levels that correlate with the smoke event.  However, no single forecast consistently predicts smoke events. Additionally, there are smoke events that none of the forecasts predict.

\subsection{Visualizing Smoke Events}
Our visualization of three separate smoke events in, respectively, three separate urban areas supports that, while there is some information in forecasts, none of the forecasts yield consistently useful predictions for individual decision making.

\textbf{Philadelphia, June 4--9, 2023:} During this event, large fires in Quebec and winds from the north led to extremely poor air quality across the northeastern United States, including in Philadelphia. Observers reported orange skies in New York City and Philadelphia on June 7, and AirNow readings exceeded 300 $\mu g/m^3$ on June 7 in Philadelphia (\cref{fig:Philadelphia-June05-09}, orange). 

\cref{fig:Philadelphia-June05-09} demonstrates that several forecasts (NAQFC, CAMS, Aurora, HAQES) fail to capture the major trends or magnitudes of pollution during this period. A user relying on these forecasts to plan across or within days would have been misled.


HRRR-Smoke accurately predicted elevated pollution levels for periods on both June 6 and 7, so an individual choosing whether to go out at all might have chosen not to. But on both days, HRRR-Smoke underestimated both the magnitude and duration of elevated PM2.5. In each case, an individual using this forecast might have planned to go outside in the evening; this person would have been exposed to much worse PM2.5 than was forecast.


GEOS-CF predicted elevated PM2.5 levels each day in June 5--7. The general forecast trend on June 6--7 was reasonably close to the observed pollution trend. The erroneous June 5 forecast might have caused individuals to take caution unnecessarily or to lose confidence in the forecast.

\begin{figure}
    \centering
    \includegraphics[width=\linewidth]{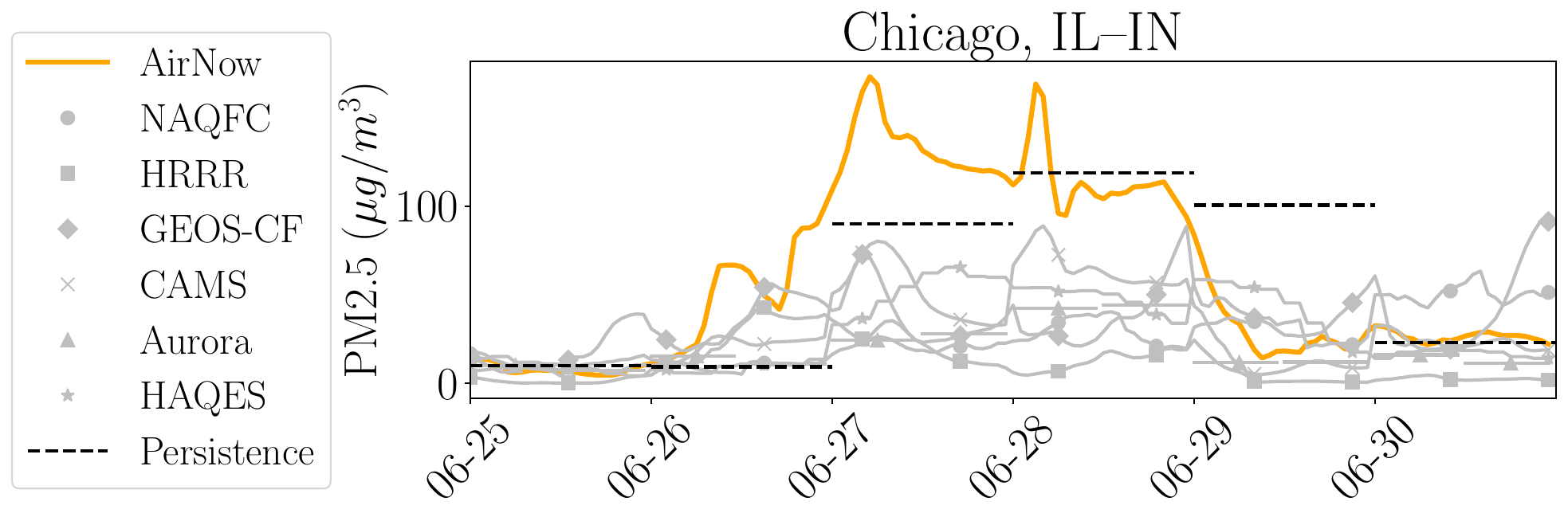}
    \caption{Air quality measurements obtained from AirNow (orange), a persistence baseline (black, dashed line) and 6 forecasts (gray, varying markers) in Chicago during June 25--30, 2023.}
    \label{fig:Chicago-June25-30}
\end{figure}

\textbf{Chicago, June 25--30, 2023:}
During this period, wind blew smoke from Canadian wildfires into Chicago, as well as the rest of the Midwestern United States. On June 27 and 28, fine particulate levels in Chicago peaked at values well in excess of 100 $\mu g/m^3$ (\cref{fig:Chicago-June25-30}, orange).

All six forecasts substantially underestimated the trend of higher PM2.5 levels during this event. So an individual planning whether to go out might have been overly optimistic. However, all forecasts predicted elevated PM2.5 levels for at least some of this period. Additionally, CAMS successfully captured the timing of the two highest peaks of PM2.5. GEOS-CF also captured the timing of the first peak but missed the timing of the second, and GEOS-CF predicted several spikes in PM2.5 levels after pollution levels had dropped substantially on June 30. An individual planning when to go out would have missed the major PM2.5 trends when using most forecasts. 

\begin{figure}
    \centering
    \includegraphics[width=\linewidth]{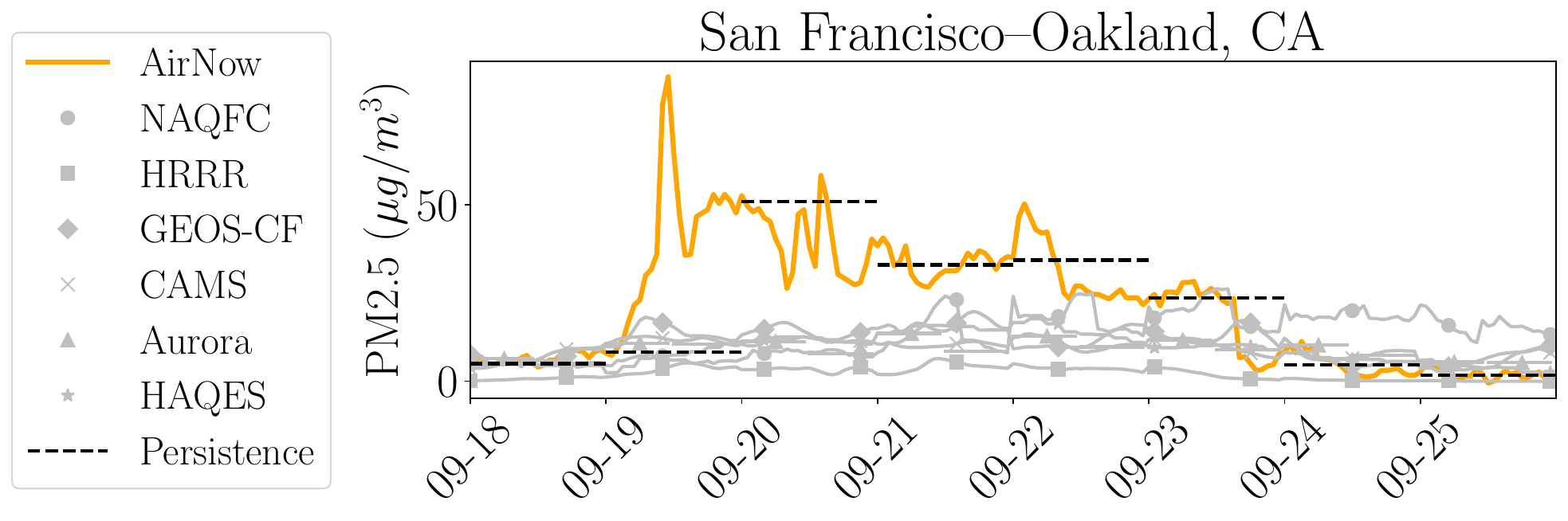}
    \caption{Air quality measurements obtained from AirNow (orange), a persistence baseline (black, dashed line) and 6 forecasts (gray, varying markers) in San Francisco during September 18--25, 2023.}
    \label{fig:SanFrancisco-Sept18-25}
\end{figure}

\textbf{San Francisco, September 18--25, 2023:}
During this period, smoke from fires in Northwestern California and Oregon led to air pollution in San Francisco and Oakland, California. The Smith River Complex fire, which burned over 94,000 acres several hundred miles north of the San Francisco Bay Area, was a primary contributor. Air pollution levels peaked in the San Francisco Area on June 19, with measured PM2.5 levels near 100 $\mu g/m^3$ (\cref{fig:SanFrancisco-Sept18-25}, orange). And PM2.5 levels remained at unhealthy levels for several days. 

No forecast predicted PM2.5 levels to exceed 35 $\mu g/m^3$, even though fine particulate pollution remained around this level for several days. 
If an individual had relied on any of these forecasts, they likely would not have taken actions to mitigate PM2.5 exposure. Meanwhile, the persistence baseline captured elevated PM2.5 levels, and officials issued air quality alerts on September 19 and 20.\footnote{\url{https://www.baaqmd.gov/news-and-events/page-resources/2023-news/091923-aq-advisory}} Since sufficient information was available for individuals to take mitigating actions, this example highlights room for improvement in all forecasts.

\input{sections/confusion_matrices}

\subsection{Identification of High-pollution Days}

Our analysis of high-pollution days across all urban areas indicates that forecasts do not reliably predict whether the coming day will have high PM2.5 levels. 

We report our results in the first six columns of \cref{tab:national-metrics}.
Of the 5,031 urban area--days we analyzed, 219 had high PM2.5 levels, and 4,812 did not achieve high PM2.5 levels. So these numbers represent, respectively, the best possible true positive and true negative values. 
Only GEOS-CF, CAMS, and HAQES have more true positives (equivalently, higher recall) than the persistence baseline. That is, the other forecasts do not identify that a day will have elevated PM2.5, even when (at the time our imagined individual consulted their forecasts) the PM2.5 level was already in excess of 35 $\mu g/m^3$. Meanwhile, both CAMS and HAQES had recall near the persistence baseline.

GEOS-CF is the only forecast with much higher recall than the persistence baseline, but it has many false positives; for every 11 days that GEOS-CF predicts smoke levels will be elevated, only one actually had smoke levels exceeding the threshold we consider. With many false positives, a user might lose confidence in the forecast.  

For an individual interested in whether the coming day will have elevated PM2.5, 
we conclude that none of the forecasts provides substantially more reliable insight than directly checking the latest AirNow readings in the morning. 
Our findings agree with \citet{ainslie_operational_2022}, who found that the Canadian FireWork forecast should not be used in isolation, without local expertise, to issue public health alerts.  We emphasize that none of these forecast systems were designed specifically for the use-case of hourly decision-making, and below we suggest ways that they might be improved.  



\subsection{Mean Excess Exposure}
Finally, we do find that forecasts can help plan when to go outside during the day.
In the second-to-last column of \cref{tab:national-metrics}, we report the MEE averaged across all days. We recognize that individuals are particularly interested in choosing when to go outside during high-pollution days (as above, where any hour exceeds 35 $\mu g/m^3$); moreover, as the climate changes, the proportion of high-pollution days is expected to change in future years. With these points in mind, we also report the average MEE across high-pollution days in the final column of \cref{tab:national-metrics}.


In this task, all forecasts (with the exception of HAQES averaged across all days) improve on the (weak) persistence baseline. The improvement on high-pollution days is substantial for the four simulation-based forecasts, with GEOS-CF, CAMS, and NAQFC nearly cutting excess exposure by half. 

\subsection{Forecast Quality in Different Regions}
We investigate whether our findings depend on the geographic region of the country. In particular, because of different climates between the eastern and western United States, it is natural to wonder if forecasts perform substantially better in one part of the country. We show in \cref{app:regional-tables} that our findings are substantively the same if we restrict our analysis to either the eastern or western United States.

%% file: sections/confusion_matrices.tex
\begin{table}[ht]
\centering
\begin{tabular}{lcccccccc}
Forecast      & TP & FP  & FN  & TN   & Precision & Recall & MEE (All Days) & MEE (Smoke Days) \\
\hline
HRRR          & 27 &   7 & 192 & 4805 & 0.794 & 0.123 & 3.52 & 12.25 \\
GEOS\textendash CF & 169 & 1892 &  50 & 2920 & 0.082 & 0.772 & 3.24 & 10.84 \\
CAMS          & 79 &  41 & 140 & 4771 & 0.658 & 0.361 & 3.10 & 10.48 \\
NAQFC         & 64 & 130 & 155 & 4682 & 0.332 & 0.292 & 3.06 & 10.82 \\
Aurora        & 25 &   7 & 194 & 4805 & 0.781 & 0.114 & 4.07 & 18.38 \\
HAQES         & 71 &  59 & 148 & 4753 & 0.546 & 0.324 & 4.55 & 17.51 \\
Persistence   & 69 &   7 & 150 & 4805 & 0.908 & 0.315 & 4.41 & 19.27 \\
\end{tabular}

\caption{For each of six forecasts and a persistence baseline (rows), we report how well each forecast identifies high PM2.5 days in the first six columns; higher is better for true positives (TP), true negatives (TN), precision, and recall --- and lower is better for false positives (FP) and false negatives (FN). In the final two columns, we report mean excess exposure (in $\mu g/m^3$) across all days and across (confirmed) high-pollution days, respectively; lower is better.}
\label{tab:national-metrics}
\end{table}

%% file: sections/appendix/setup_details.tex
\section{Details of Scope of the Study}

\subsection{A List of Cities and States Characterizing the Urban Areas We Consider}\label{app:metro-areas}

\begin{itemize}
    \item \textbf{New England:}
        \begin{itemize}
            \item Boston, MA--NH
            \item Worcester, MA--CT
        \end{itemize}
    \item \textbf{Middle Atlantic:}
        \begin{itemize}
            \item New York--Jersey City--Newark, NY--NJ
            \item Philadelphia, PA--NJ--DE--MD
        \end{itemize}
    \item \textbf{East North Central:}
        \begin{itemize}
            \item Chicago, IL--IN
            \item Detroit, MI
        \end{itemize}
    \item \textbf{West North Central:}
        \begin{itemize}
            \item Minneapolis--St.\ Paul, MN
            \item St.\ Louis, MO--IL
        \end{itemize}
    \item \textbf{South Atlantic:}
        \begin{itemize}
            \item Washington--Arlington, DC--VA--MD
            \item Atlanta, GA
            \item Miami--Fort Lauderdale, FL
            \item Baltimore, MD
            \item Orlando, FL
            \item Charlotte, NC--SC
        \end{itemize}
    \item \textbf{East South Central:}
        \begin{itemize}
            \item Nashville--Davidson, TN
            \item Memphis, TN--MS--AR
            \item Louisville--Jefferson County, KY--IN
        \end{itemize}
    \item \textbf{West South Central:}
        \begin{itemize}
            \item Dallas--Fort Worth--Arlington, TX
            \item Houston, TX
            \item San Antonio, TX
        \end{itemize}
    \item \textbf{Mountain:}
        \begin{itemize}
            \item Phoenix--Mesa--Scottsdale, AZ
            \item Denver--Aurora, CO 
            \item Las Vegas--Henderson--Paradise, NV
        \end{itemize}
    \item \textbf{Pacific:}
        \begin{itemize}
            \item Los Angeles--Long Beach--Anaheim, CA
            \item San Francisco--Oakland, CA
            \item Riverside--San Bernardino, CA 
            \item Seattle--Tacoma, WA 
            \item Portland, OR--WA 
            \item San Diego, CA
            \item Tampa--St.\ Petersburg, FL
        \end{itemize}
\end{itemize}

\subsection{Spatial Averaging}\label{app:spatial-averaging}
Forecast predictions are generally not available at the exact locations of PM2.5 monitors. To compute the ``ground truth'' PM2.5 for an urban area, we use latitude and longitude data provided by the US Census Gazetteer list of Urban Areas \citep{census}. We perform a nearest-neighbors search using the Haversine distance to find the nearest active monitors within a 10 km radius of each urban area, with a maximum of 10 neighbors included. If the urban areas had no PM2.5 monitors within a 10 km radius of the latitude and longitude provided in the gazetteer, we expanded the radius of monitors included to 50 km. We then calculate PM2.5 concentration for the urban area as the average of the PM2.5 concentrations measured by all included monitors.

We similarly calculate the forecast PM2.5 in an urban area as a spatial average of monitors close to the urban area. We average all predictions within a fixed radius of the latitude and longitude used for the urban area. We use a radius of 50 km for HRRR-Smoke, GEOS-CF, HAQES, Aurora, and NAQFC and a radius of 60 km for CAMS due to its coarser spatial resolution. 

We perform spatial averaging to (1) reduce potential variability in sensor readings due to imperfect measurements and (2) account for the urban areas involved being regions and not points. Forecasts are often considered at a regional level, and so considering a spatial region instead of a single point can be useful in defining evaluation metrics \citep{NewCategoricalMetricsforAirQualityModelEvaluation}.

\subsection{Selection of Forecast Cycles}\label{app:forecast-timings}

\textbf{High Resolution Rapid-Refresh Smoke (HRRR-Smoke).} 

 We use the cycle of HRRR-Smoke run at 12UTC. We compare forecasts over a 24-hour window beginning at 13UTC and running until 12UTC the following day.
 
\textbf{Copernicus Atmosphere
Monitoring Service Forecast (CAMS).}
 We use the cycle of CAMS run at 12UTC. We compare forecasts over a 24-hour window beginning at 13UTC and running until 12UTC the following day.
 
\textbf{GEOS Composition Forecasting (GEOS-CF).} 
 We use the cycle of GEOS-CF run at 12UTC. We compare forecasts over a 24-hour window beginning at 13UTC and running until 12UTC the following day. GEOS-CF computes both time-averaged (over an hour) and instantaneous forecast products. We use the time-averaged version, and we associate the forecast to the earlier hour. For instance, the forecast between 12UTC and 13UTC is associated to 12UTC.
 
\textbf{National Air Quality Forecast Capability (NAQFC).}

 We use the cycle of NAQFC run at 12UTC. We compare forecasts over a 24-hour window beginning at 13UTC and running until 12UTC the following day.
 
\textbf{Hazardous Air Quality Ensemble System (HAQES).}

HAQES is run on a single cycle at 12UTC, and makes predictions from 0UTC--24UTC the following day. We always use the cycle run on the prior date (UTC). It follows that, for each day (which for us ranges from 13UTC to 12UTC+1), we use distinct forecasts for the first and second 12 hours. HAQES makes predictions at a 3-hour time resolution. We treat these predictions as constant for the 3 hours prior to each prediction to handle the coarser time resolution of the forecast. 

\textbf{Microsoft Aurora.}
The Aurora model was fine-tuned to make 12 hour predictions on a twice daily cycle. We use prediction made at 12UTC for 0UTC and 12UTC the following day. We treat these predictions as constant over the 12 hours prior to the prediction to handle the coarser time resolution of the forecast. 

\textbf{Persistence Baseline.} For the persistence baseline we use the monitor reading from 11UTC, and treat it as a forecast predicting that value forward from 13UTC on the day of the reading until 12UTC the following day. 

\subsection{Regional Results Tables}
\label{app:regional-tables}
\input{sections/appendix/regional_tables}

We next confirm that we see similar performs patterns separately in the Eastern and Western Regions of the continental US.
In particular, no forecast performance substantially better than the persistence baseline for the task of deciding whether to go outside --- although CAMS stands out more for desirable performance in the Eastern Region. And multiple forecasts (especially GEOS-CF, CAMS, and NAQFC) are substantively better than the persistence baseline at deciding when to go outside. 

In \cref{tab:eastern_results,tab:western_results}, we report the same metrics as in \cref{tab:national-metrics}, but now for the Eastern Region and Western Region separately.

The Eastern Region consists of 21 U.S.\ urban areas, namely those listed in \cref{app:metro-areas} and corresponding to: Atlanta, GA; Baltimore, MD; Boston, MA; Charlotte, NC; Chicago, IL; Dallas, TX; Detroit, MI; Houston, TX; Louisville, KY; Memphis, TN; Miami, FL; Minneapolis, MN; Nashville, TN; New York, NY; Orlando, FL; Philadelphia, PA; San Antonio, TX; St.~Louis, MO; Tampa, FL; Washington, DC; Worcester, MA.

The Western Region consists of 9 U.S.\ urban areas, namely those corresponding to: Denver, CO; Las Vegas, NV; Los Angeles, CA; Phoenix, AZ; Portland, OR; Riverside, CA; San Diego, CA; San Francisco, CA; Seattle, WA.

Of the total 219 total high-pollution urban area--days, 182 are in the Eastern Region, and 37 are in the Western Region. The Eastern Region had 3247 urban area--days without high pollution, and the Western Region had 1565 without high pollution.

\newpage

%% file: sections/appendix/regional_tables.tex
\begin{table}[ht]
\centering
\caption{The rows and columns are as in \cref{tab:national-metrics}, but now we report all values only over the \textbf{Eastern Region}.}
\begin{tabular}{lcccccccc}
Forecast & TP & FP & FN & TN & Precision & Recall & MEE (All Days) & MEE (Smoke Days) \\
\hline
HRRR          & 24 &   4 & 158 & 3243 & 0.857 & 0.132 & 3.79 & 11.65 \\
Geos-CF & 147 & 1499 &  35 & 1748 & 0.089 & 0.808 & 3.73 & 11.76 \\
CAMS          & 75 &  37 & 107 & 3210 & 0.670 & 0.412 & 3.29 & 10.11 \\
NAQFC         & 57 & 129 & 125 & 3118 & 0.306 & 0.313 & 3.42 & 11.50 \\
Aurora        & 24 &   7 & 158 & 3240 & 0.774 & 0.132 & 4.44 & 19.49 \\
HAQES         & 67 &  54 & 115 & 3193 & 0.554 & 0.368 & 4.93 & 17.45 \\
Persistence   & 61 &   7 & 121 & 3240 & 0.897 & 0.335 & 4.90 & 20.15 \\
\end{tabular}
\label{tab:eastern_results}
\end{table}

\begin{table}[ht]
\centering
\caption{The rows and columns are as in \cref{tab:national-metrics}, but now we report all values only over the \textbf{Western Region}.}
\begin{tabular}{lcccccccc}
Forecast & TP & FP & FN & TN & Precision & Recall & MEE (All Days) & MEE (Smoke Days) \\
\hline
HRRR          &  3 &   3 & 34 & 1562 & 0.500 & 0.081 & 2.95 & 15.23 \\
Geos-CF & 22 & 393 & 15 & 1172 & 0.053 & 0.595 & 2.18 &  6.32 \\
CAMS          &  4 &   4 & 33 & 1561 & 0.500 & 0.108 & 2.67 & 12.30 \\
NAQFC         &  7 &   1 & 30 & 1564 & 0.875 & 0.189 & 2.29 &  7.48 \\
Aurora        &  1 &   0 & 36 & 1565 & 1.000 & 0.027 & 3.28 & 12.93 \\
HAQES         &  4 &   5 & 33 & 1560 & 0.444 & 0.108 & 3.75 & 17.82 \\
Persistence   &  8 &   0 & 29 & 1565 & 1.000 & 0.216 & 3.37 & 14.93 \\
\end{tabular}
\label{tab:western_results}
\end{table}

%% file: sections/supplement/forecasts.tex
We consider four forecasts based on meteorological and chemical simulation, one ensemble of forecasts, a machine learning forecast and a persistence baseline. When multiple versions of a forecast exist, we focus on the latest version of the forecast that was active during the 2023 fire season.

\section{High-Resolution Rapid Refresh Smoke (HRRRv4-Smoke)} 

The National Oceanic and Atmospheric Administration (NOAA) develops and maintains High-Resolution Rapid Refresh (HRRR) \citep{dowell2022hrrrsmoke}.  HRRR is a numerical weather prediction model. The NWS released version 4 of HRRR, which we use, in 2020. HRRR operates on 3km resolution over the extent of the contiguous United States and Alaska. HRRR produces 18 hour forecasts for the continental United States every hour and 48 hour forecasts every 6 hours. To calculate PM2.5 sources, HRRR-Smoke relies on fire radiative power (FRP) estimates obtained from remote sensing (VIIRS and MODIS) data \citet{Ahmadov2017VIIRSFRP}. HRRR estimates the smoke plume based on FRP together with information about the land on which the fire is burning. HRRR integrates smoke transport using a numerical weather prediction model and a simplified version of WRF-chem \citet{grell2005wrfchem}. As HRRR-Smoke accounts for only smoke sources that remote sensing can detect (wildfires and prescribed burns), we expect it to systematically underestimate PM2.5 levels, particularly in locations where there are significant anthropogenic sources of fine particulate matter. However, because wildfire sources drive many of the extreme PM2.5 events in the United States,  HRRR-Smoke is a plausible tool to predict the intensity and timing of extremely elevated fine particulate matter events in the continental United States.

\section{Copernicus Atmosphere
Monitoring Service Forecast (CAMS, cycle 47r3 (May 1st--June 27) cycle 48r1 (June 28--October 1st)} The Copernicus Atmosphere Monitoring Service (CAMS) is an air quality and atmospheric composition forecast system the European Centre for Medium-Range Weather Forecasts (ECMWF) developed as part of the ECMWF Integrated Forecast System \citep{ecmwf2023cams}. We use the operational version of the forecast on each day; the ECWMF updated CAMS on June 27th, from cycle 47r3 to cycle 48r1\footnote{https://www.ecmwf.int/en/forecasts/documentation-and-support/changes-ecmwf-model}.  CAMS provides global and regional forecasts of atmospheric composition, including concentrations of key pollutants such as particulate matter (PM2.5), ozone, and nitrogen dioxide. The model operates at an effective horizontal resolution of 0.4°, approximately 40 km, making it suitable for broad-scale analysis over large regions. CAMS integrates data from multiple sources, including satellite observations and ground-based measurements, to produce air quality forecasts. CAMS uses remote sensing estimates of FRP to estimate biomass burning emissions.

\section{GEOS Composition Forecasting (GEOS-CFv1)} The NASA's Global Modeling and Assimilation Office (GMAO)  develops and maintains the Global Earth Observing System Composition Forecasting (GEOS-CF) model \citep{keller2021geoscf}. We look at forecasts from GEOS-CFv1, as this was the active version of GEOS-CF in 2023; NASA plans GEOS-CFv2 in the future \citep{Knowland2024Geoscfv2}. GEOS-CF provides global air quality forecasts, leveraging the GEOS Atmospheric General Circulation Model and a sophisticated data assimilation system (hybrid-4DEnVar ADAS) \citep{gelaro2017modern}. The model operates at a spatial resolution of 0.25° (approximately 25 km) and includes 72 vertical layers. GEOS-CF runs 5-day forecasts once per day, offering both high temporal resolution products with outputs every 15 minutes or hourly, which can be either time-averaged or instantaneous. The model integrates a wide range of emission sources, including near-real-time biomass burning emissions from the Quick Fire Emission Database (QFED v2.5) \citep{darmenov2015quick} and anthropogenic emissions from the HTAP v2.2 and RETRO inventories \citep{janssens2015htap_v2, schultz2008global}, which are broken down into hourly values using sector-specific day-of-week and diurnal scale factors \citep{denier2013policy}. The GOCART model simulates aerosol components \citep{chin2002tropospheric}, incorporating both anthropogenic and biogenic emissions.

\section{National Air Quality Forecast Capability (NAQFC, AQMv6)} NOAA and the NWS developed and maintain the National Air Quality Forecast Capability (NAQFC) \citep{davidson2008naqfc,stajner2012naqfcpm}. NAQFC integrates the North American Mesoscale (NAM) Forecast System with the Community Multiscale Air Quality (CMAQ) modeling system. NAQFC provides air quality forecasts up to 48 hours out, issued twice daily at 6 and 12 UTC, covering the contiguous United States, Alaska, and Hawaii at a spatial resolution of approximately 12 km. The model operates using NOAA’s operational FV3-GFSv16 meteorology and includes 35 vertical layers, enabling detailed simulations of atmospheric processes \citep{campbell2022development}. NAQFC employs static chemical gaseous boundary conditions from global GEOS-Chem simulations, while aerosol boundary conditions are dynamically updated from NOAA’s operational GEFS-Aerosols model \citep{zhang2022development}. Biomass-burning emissions are incorporated into the model via the GBBEPx system, with wildfire smoke plumes computed using the Briggs plume rise algorithm \citep{briggs1975plume}. In addition, NAQFC accounts for anthropogenic and biogenic emissions, providing a comprehensive representation of various emission sources that affect air quality. Fire emissions are integrated into PM2.5 concentration forecasts using the BlueSky framework \citep{larkin2009bluesky}. NAQFC's capability to forecast PM2.5 was released in 2015/2016, much later than its ozone forecasts, and has undergone continuous improvements to better capture the effects of fires and extreme pollution events. Despite its robust framework, the model faces challenges in accurately predicting air quality during extreme events, as noted in NOAA’s documentation \citep{naqfc2016report}. To address some of these challenges, \citet{huang2017improving} considered bias correction approaches to improve NAQFC's predictions of PM2.5 levels during extreme events. Because we look at data from 2023, we consider forecasts made by AQMv6, which was the operational forecast during 2023. NOAA released AQMv7 \citep{DevelopmentofthenextgenerationairqualitypredictionsystemintheUnifiedForecastSystemframeworkEnhancingpredictabilityofwildfireairqualityimpacts} in 2024.

\section{Hazardous Air Quality Ensemble System (HAQES)}
The Hazardous Air Quality Ensemble System (HAQES) \citep{MultiagencyEnsembleForecastofWildfireAirQualityintheUnitedStatesTowardCommunityConsensusofEarlyWarning} combines predictions from HRRR-Smoke \citep{dowell2022hrrrsmoke}, GEFS-Aerosols \citep{zhang2022development}, NAQFC \citep{campbell2022development}, GEOS-FP \citep{gelaro2017modern}, and NAAPS \citep{CHRISTENSEN19974169,gmd-9-1489-2016}. \Citet{tong2023haqes} reports the mean of the ensemble members. A single forecast is made daily, using forecasts initialized at 12UTC the previous day. The forecasts ranges from 0UTC to 23UTC for the current day, at a 3 hour time resolution. The spatial resolution is $12$km $\times 12$km.

\citet{MultiagencyEnsembleForecastofWildfireAirQualityintheUnitedStatesTowardCommunityConsensusofEarlyWarning} showed that a weighted least squares approach, with more importance given to extreme events, leads to ensemble weights with improved forecasting of which days will have elevated PM2.5 levels. However, to our knowledge, the weighting approach they proposed has not been made into a publicly available forecast.

\section{Microsoft Aurora}
Aurora is a state-of-the-art machine learning model for weather forecasting \citep{bodnar_foundation_2025}. The training process for Aurora follows a similar paradigm to training of many modern artificial intelligence systems, especially large language models. In particular, Microsoft initially trained Aurora on a large amount of earth system analysis and reanalysis datasets, and subsequently Microsoft ``fine-tuned” Aurora specifically for air quality forecasting using CAMS analysis and reanalysis data. However, this fine-tuning step is constrained by the quality and comprehensiveness of the information in the CAMS data. Currently, Aurora incorporates wildfire information solely through CAMS analysis, meaning that important physical processes --- such as fire emissions, plume rise, and transport dynamics --- are only indirectly represented, rather than explicitly integrated into the model. Aurora makes predictions at 12 hour time resolution, and a spatial resolution of $0.25^{\circ}$.

\section*{Persistence Baseline} 
\citet{ainslie_operational_2022} advocate for a baseline forecast that predicts air quality will remain constant for the next 24 hours. We construct this baseline from monitor data publicly available through the EPA AirNow system \citep{airnow2024}.

%% file: main.bbl
\begin{thebibliography}{42}
\providecommand{\natexlab}[1]{#1}
\providecommand{\url}[1]{\texttt{#1}}
\expandafter\ifx\csname urlstyle\endcsname\relax
  \providecommand{\doi}[1]{doi: #1}\else
  \providecommand{\doi}{doi: \begingroup \urlstyle{rm}\Url}\fi

\bibitem[Abatzoglou and Williams(2016)]{abatzoglou2016impact}
John~T Abatzoglou and A~Park Williams.
\newblock Impact of anthropogenic climate change on wildfire across western {US} forests.
\newblock \emph{Proceedings of the National Academy of sciences}, 113\penalty0 (42):\penalty0 11770--11775, 2016.

\bibitem[Adetona et~al.(2016)Adetona, ~, ~, ~, ~, ~, ~, , and Naeher]{adetona_review_2016}
Olorunfemi Adetona, Reinhardt ~, Timothy~E., Domitrovich ~, Joe, Broyles ~, George, Adetona ~, Anna~M., Kleinman ~, Michael~T., Ottmar ~, Roger~D., , and Luke~P. Naeher.
\newblock Review of the health effects of wildland fire smoke on wildland firefighters and the public.
\newblock \emph{Inhalation Toxicology}, 28\penalty0 (3):\penalty0 95--139, February 2016.

\bibitem[Ahmadov et~al.(2017)Ahmadov, Grell, James, Csiszar, Tsidulko, Pierce, McKeen, Benjamin, Alexander, Pereira, Freitas, and Goldberg]{Ahmadov2017VIIRSFRP}
R.~Ahmadov, G.~Grell, E.~James, I.~Csiszar, M.~Tsidulko, B.~Pierce, S.~McKeen, S.~Benjamin, C.~Alexander, G.~Pereira, S.~Freitas, and M.~Goldberg.
\newblock Using {VIIRS} fire radiative power data to simulate biomass burning emissions, plume rise and smoke transport in a real-time air quality modeling system.
\newblock In \emph{2017 IEEE International Geoscience and Remote Sensing Symposium (IGARSS)}, pages 2806--2808, 2017.
\newblock \doi{10.1109/IGARSS.2017.8127581}.

\bibitem[Ainslie et~al.(2022)Ainslie, So, and Chen]{ainslie_operational_2022}
Bruce Ainslie, Rita So, and Jack Chen.
\newblock Operational {Evaluation} of a {Wildfire} {Air} {Quality} {Model} from a {Forecaster} {Point} of {View}.
\newblock \emph{Weather and Forecasting}, 37\penalty0 (5):\penalty0 681--698, May 2022.

\bibitem[Bodnar et~al.(2025)Bodnar, Bruinsma, Lucic, Stanley, Allen, Brandstetter, Garvan, Riechert, Weyn, Dong, Gupta, Thambiratnam, Archibald, Wu, Heider, Welling, Turner, and Perdikaris]{bodnar_foundation_2025}
Cristian Bodnar, Wessel~P. Bruinsma, Ana Lucic, Megan Stanley, Anna Allen, Johannes Brandstetter, Patrick Garvan, Maik Riechert, Jonathan~A. Weyn, Haiyu Dong, Jayesh~K. Gupta, Kit Thambiratnam, Alexander~T. Archibald, Chun-Chieh Wu, Elizabeth Heider, Max Welling, Richard~E. Turner, and Paris Perdikaris.
\newblock A foundation model for the {Earth} system.
\newblock \emph{Nature}, 641\penalty0 (8065):\penalty0 1180--1187, May 2025.

\bibitem[Briggs(1975)]{briggs1975plume}
Gary~A Briggs.
\newblock Plume rise predictions.
\newblock In \emph{Lectures on air pollution and environmental impact analyses}, pages 59--111. Springer, 1975.

\bibitem[Burke et~al.(2023)Burke, Childs, de~la Cuesta, Qiu, Li, Gould, Heft-Neal, and Wara]{burke_contribution_2023}
Marshall Burke, Marissa~L. Childs, Brandon de~la Cuesta, Minghao Qiu, Jessica Li, Carlos~F. Gould, Sam Heft-Neal, and Michael Wara.
\newblock The contribution of wildfire to {PM2}.5 trends in the {USA}.
\newblock \emph{Nature}, 622\penalty0 (7984):\penalty0 761--766, October 2023.

\bibitem[Campbell et~al.(2022)Campbell, Tang, Lee, Baker, Tong, Saylor, Stein, Huang, Huang, Strobach, et~al.]{campbell2022development}
Patrick~C Campbell, Youhua Tang, Pius Lee, Barry Baker, Daniel Tong, Rick Saylor, Ariel Stein, Jianping Huang, Ho-Chun Huang, Edward Strobach, et~al.
\newblock Development and evaluation of an advanced national air quality forecasting capability using the {NOAA} global forecast system version 16.
\newblock \emph{Geoscientific model development}, 15\penalty0 (8):\penalty0 3281--3313, 2022.

\bibitem[Chen et~al.(2019)Chen, Anderson, Pavlovic, Moran, Englefield, Thompson, Munoz-Alpizar, and Landry]{chen2019firework}
J.~Chen, K.~Anderson, R.~Pavlovic, M.~D. Moran, P.~Englefield, D.~K. Thompson, R.~Munoz-Alpizar, and H.~Landry.
\newblock The {FireWork v2.0} air quality forecast system with biomass burning emissions from the canadian forest fire emissions prediction system {v2.03}.
\newblock \emph{Geoscientific Model Development}, 12\penalty0 (7):\penalty0 3283--3310, 2019.

\bibitem[Chin et~al.(2002)Chin, Ginoux, Kinne, Torres, Holben, Duncan, Martin, Logan, Higurashi, and Nakajima]{chin2002tropospheric}
Mian Chin, Paul Ginoux, Stefan Kinne, Omar Torres, Brent~N Holben, Bryan~N Duncan, Randall~V Martin, Jennifer~A Logan, Akiko Higurashi, and Teruyuki Nakajima.
\newblock Tropospheric aerosol optical thickness from the {GOCART} model and comparisons with satellite and sun photometer measurements.
\newblock \emph{Journal of the atmospheric sciences}, 59\penalty0 (3):\penalty0 461--483, 2002.

\bibitem[Chow et~al.(2022)Chow, Yu, Young, James, Grell, Csiszar, Tsidulko, Freitas, Pereira, Giglio, Friberg, and Ahmadov]{chow_high-resolution_2022}
Fotini~Katopodes Chow, Katelyn~A. Yu, Alexander Young, Eric James, Georg~A. Grell, Ivan Csiszar, Marina Tsidulko, Saulo Freitas, Gabriel Pereira, Louis Giglio, Mariel~D. Friberg, and Ravan Ahmadov.
\newblock High-resolution smoke forecasting for the 2018 {Camp} {Fire} in {California}.
\newblock \emph{Bulletin of the American Meteorological Society}, 103\penalty0 (6):\penalty0 E1531--E1552, June 2022.

\bibitem[Christensen(1997)]{CHRISTENSEN19974169}
Jesper~Heile Christensen.
\newblock The {D}anish {E}ulerian hemispheric model --- a three-dimensional air pollution model used for the arctic.
\newblock \emph{Atmospheric Environment}, 31\penalty0 (24):\penalty0 4169--4191, 1997.

\bibitem[Darmenov and da~Silva(2015)]{darmenov2015quick}
Anton Darmenov and Arlindo da~Silva.
\newblock The quick fire emissions dataset ({QFED}): Documentation of versions 2.1, 2.2 and 2.4.
\newblock \emph{NASA Technical Report Series on Global Modeling and Data Assimilation, NASA TM-2013-104606}, 32:\penalty0 183, 2015.

\bibitem[Davidson et~al.(2008)Davidson, Schere, Draxler, Kondragunta, Wayland, Meagher, and Mathur]{davidson2008naqfc}
Paula Davidson, Kenneth Schere, Roland Draxler, Shobha Kondragunta, Richard~A. Wayland, James~F. Meagher, and Rohit Mathur.
\newblock Toward a {US} national air quality forecast capability: Current and planned capabilities.
\newblock In Carlos Borrego and Ana~Isabel Miranda, editors, \emph{Air Pollution Modeling and Its Application XIX}, pages 226--234. Springer Netherlands, 2008.

\bibitem[DeFlorio-Barker et~al.(2019)DeFlorio-Barker, Crooks, Reyes, and Rappold]{deflorio-barker_cardiopulmonary_2019}
Stephanie DeFlorio-Barker, James Crooks, Jeanette Reyes, and Ana~G. Rappold.
\newblock Cardiopulmonary effects of fine particulate matter exposure among older {Adults}, during wildfire and non-wildfire periods, in the {United} {States} 2008-2010.
\newblock \emph{Environmental Health Perspectives}, 127\penalty0 (3):\penalty0 37006, March 2019.

\bibitem[Denier van~der Gon et~al.(2013)Denier van~der Gon, Gerlofs-Nijland, Gehrig, Gustafsson, Janssen, Harrison, Hulskotte, Johansson, Jozwicka, Keuken, et~al.]{denier2013policy}
Hugo~AC Denier van~der Gon, Miriam~E Gerlofs-Nijland, Robert Gehrig, Mats Gustafsson, Nicole Janssen, Roy~M Harrison, Jan Hulskotte, Christer Johansson, Magdalena Jozwicka, Menno Keuken, et~al.
\newblock The policy relevance of wear emissions from road transport, now and in the future—an international workshop report and consensus statement.
\newblock \emph{Journal of the Air \& Waste Management Association}, 63\penalty0 (2):\penalty0 136--149, 2013.

\bibitem[Dominici et~al.(2006)Dominici, Peng, Bell, Pham, McDermott, Zeger, and Samet]{Dominici2006Fine}
Francesca Dominici, Roger~D. Peng, Michelle~L. Bell, Luu Pham, Aidan McDermott, Scott~L. Zeger, and Jonathan~M. Samet.
\newblock Fine particulate air pollution and hospital admission for cardiovascular and respiratory diseases.
\newblock \emph{JAMA}, 295\penalty0 (10):\penalty0 1127--1134, 03 2006.
\newblock ISSN 0098-7484.
\newblock \doi{10.1001/jama.295.10.1127}.
\newblock URL \url{https://doi.org/10.1001/jama.295.10.1127}.

\bibitem[Dowell et~al.(2022)Dowell, Alexander, James, Weygandt, Benjamin, Manikin, Blake, Brown, Olson, Hu, Smirnova, Ladwig, Kenyon, Ahmadov, Turner, Duda, and Alcott]{dowell2022hrrrsmoke}
David~C. Dowell, Curtis~R. Alexander, Eric~P. James, Stephen~S. Weygandt, Stanley~G. Benjamin, Geoffrey~S. Manikin, Benjamin~T. Blake, John~M. Brown, Joseph~B. Olson, Ming Hu, Tatiana~G. Smirnova, Terra Ladwig, Jaymes~S. Kenyon, Ravan Ahmadov, David~D. Turner, Jeffrey~D. Duda, and Trevor~I. Alcott.
\newblock The high-resolution rapid refresh ({HRRR}): An hourly updating convection-allowing forecast model.{Part I}: Motivation and system description.
\newblock \emph{Weather and Forecasting}, 37\penalty0 (8):\penalty0 1371 -- 1395, 2022.

\bibitem[{European Centre for Medium-Range Weather Forecasts}(2023)]{ecmwf2023cams}
{European Centre for Medium-Range Weather Forecasts}.
\newblock \emph{IFS Documentation CY48R1 - Part VIII: Atmospheric Composition}.
\newblock Number~8. ECMWF, 6 2023.
\newblock \doi{10.21957/749dc09059}.

\bibitem[Gelaro et~al.(2017)Gelaro, McCarty, Su{\'a}rez, Todling, Molod, Takacs, Randles, Darmenov, Bosilovich, Reichle, et~al.]{gelaro2017modern}
Ronald Gelaro, Will McCarty, Max~J Su{\'a}rez, Ricardo Todling, Andrea Molod, Lawrence Takacs, Cynthia~A Randles, Anton Darmenov, Michael~G Bosilovich, Rolf Reichle, et~al.
\newblock The modern-era retrospective analysis for research and applications, version 2 ({MERRA}-2).
\newblock \emph{Journal of climate}, 30\penalty0 (14):\penalty0 5419--5454, 2017.

\bibitem[Grell et~al.(2005)Grell, Peckham, Schmitz, McKeen, Frost, Skamarock, and Eder]{grell2005wrfchem}
Georg~A. Grell, Steven~E. Peckham, Rainer Schmitz, Stuart~A. McKeen, Gregory Frost, William~C. Skamarock, and Brian Eder.
\newblock Fully coupled ``online'' chemistry within the {WRF} model.
\newblock \emph{Atmospheric Environment}, 39\penalty0 (37):\penalty0 6957--6975, 2005.

\bibitem[Holstius et~al.(2012)Holstius, Reid, Jesdale, and Morello-Frosch]{holstius_birth_2012}
David~M. Holstius, Colleen~E. Reid, Bill~M. Jesdale, and Rachel Morello-Frosch.
\newblock Birth weight following pregnancy during the 2003 {Southern} {California} wildfires.
\newblock \emph{Environmental Health Perspectives}, 120\penalty0 (9):\penalty0 1340--1345, September 2012.

\bibitem[Huang et~al.(2017)Huang, McQueen, Wilczak, Djalalova, Stajner, Shafran, Allured, Lee, Pan, Tong, et~al.]{huang2017improving}
Jianping Huang, Jeffery McQueen, James Wilczak, Irina Djalalova, Ivanka Stajner, Perry Shafran, Dave Allured, Pius Lee, Li~Pan, Daniel Tong, et~al.
\newblock Improving {NOAA NAQFC PM2.5} predictions with a bias correction approach.
\newblock \emph{Weather and Forecasting}, 32\penalty0 (2):\penalty0 407--421, 2017.

\bibitem[Huang et~al.(2025)Huang, Stajner, Montuoro, Yang, Wang, Huang, Jeon, Curtis, McQueen, Liu, Baker, Tong, Tang, Campbell, Grell, Frost, Schwantes, Wang, Kondragunta, Li, and Jung]{DevelopmentofthenextgenerationairqualitypredictionsystemintheUnifiedForecastSystemframeworkEnhancingpredictabilityofwildfireairqualityimpacts}
Jianping Huang, Ivanka Stajner, Raffaele Montuoro, Fanglin Yang, Kai Wang, Ho-Chun Huang, Chan-Hoo Jeon, Brian Curtis, Jeff McQueen, Haixia Liu, Barry Baker, Daniel Tong, Youhua Tang, Patrick Campbell, Georg Grell, Gregory Frost, Rebecca Schwantes, Siyuan Wang, Shobha Kondragunta, Fangjun Li, and Youngsun Jung.
\newblock Development of the next-generation air quality prediction system in the unified forecast system framework: Enhancing predictability of wildfire air quality impacts.
\newblock \emph{Bulletin of the American Meteorological Society}, 2025.

\bibitem[Janssens-Maenhout et~al.(2015)Janssens-Maenhout, Crippa, Guizzardi, Dentener, Muntean, Pouliot, Keating, Zhang, Kurokawa, Wankm{\"u}ller, et~al.]{janssens2015htap_v2}
G~Janssens-Maenhout, Monica Crippa, Diego Guizzardi, Frank Dentener, Marilena Muntean, George Pouliot, Terry Keating, Qiang Zhang, Junishi Kurokawa, Robert Wankm{\"u}ller, et~al.
\newblock {HTAP\_v2}. 2: a mosaic of regional and global emission grid maps for 2008 and 2010 to study hemispheric transport of air pollution.
\newblock \emph{Atmospheric Chemistry and Physics}, 15\penalty0 (19):\penalty0 11411--11432, 2015.

\bibitem[Kang et~al.(2007)Kang, Mathur, Schere, Yu, and Eder]{NewCategoricalMetricsforAirQualityModelEvaluation}
Daiwen Kang, Rohit Mathur, Kenneth Schere, Shaocai Yu, and Brian Eder.
\newblock New categorical metrics for air quality model evaluation.
\newblock \emph{Journal of Applied Meteorology and Climatology}, 46\penalty0 (4):\penalty0 549 -- 555, 2007.

\bibitem[Keller et~al.(2021)Keller, Knowland, Duncan, Liu, Anderson, Das, Lucchesi, Lundgren, Nicely, Nielsen, Ott, Saunders, Strode, Wales, Jacob, and Pawson]{keller2021geoscf}
Christoph~A. Keller, K.~Emma Knowland, Bryan~N. Duncan, Junhua Liu, Daniel~C. Anderson, Sampa Das, Robert~A. Lucchesi, Elizabeth~W. Lundgren, Julie~M. Nicely, Eric Nielsen, Lesley~E. Ott, Emily Saunders, Sarah~A. Strode, Pamela~A. Wales, Daniel~J. Jacob, and Steven Pawson.
\newblock Description of the {NASA GEOS} composition forecast modeling system {GEOS-CF v1.0}.
\newblock \emph{Journal of Advances in Modeling Earth Systems}, 13\penalty0 (4):\penalty0 e2020MS002413, 2021.

\bibitem[Knowland et~al.(2024)Knowland, Keller, Shah, and Wales]{Knowland2024Geoscfv2}
K.~Emma Knowland, Christoph~A. Keller, Viral Shah, and Pam Wales.
\newblock {NASA GEOS} composition forecast system, {GEOS-CF}: {TEMPO} support, 2024.
\newblock URL \url{https://haqast.org/wp-content/uploads/sites/91/2024/10/poster_GEOS-CF_HAQAST_June2024.pdf}.

\bibitem[Larkin et~al.(2009)Larkin, O’Neill, Solomon, Raffuse, Strand, Sullivan, Krull, Rorig, Peterson, and Ferguson]{larkin2009bluesky}
Narasimhan~K Larkin, Susan~M O’Neill, Robert Solomon, Sean Raffuse, Tara Strand, Dana~C Sullivan, Candace Krull, Miriam Rorig, Janice Peterson, and Sue~A Ferguson.
\newblock The {BlueSky} smoke modeling framework.
\newblock \emph{International journal of wildland fire}, 18\penalty0 (8):\penalty0 906--920, 2009.

\bibitem[Lee et~al.(2024)Lee, Alessandrini, Kim, Meech, Kumar, Djalalova, and Wilczak]{LEE2024120833}
Jared~A. Lee, Stefano Alessandrini, Ju-Hye Kim, Scott Meech, Rajesh Kumar, Irina~V. Djalalova, and James~M. Wilczak.
\newblock Comparison of cams and cmaq analyses of surface-level pm2.5 and o3 over the conterminous united states (conus).
\newblock \emph{Atmospheric Environment}, 338:\penalty0 120833, 2024.

\bibitem[Li et~al.(2024)Li, Tong, Makkaroon, DelSole, Tang, Campbell, Baker, Cohen, Darmenov, Ahmadov, James, Hyer, and Xian]{MultiagencyEnsembleForecastofWildfireAirQualityintheUnitedStatesTowardCommunityConsensusofEarlyWarning}
Yunyao Li, Daniel Tong, Peewara Makkaroon, Timothy DelSole, Youhua Tang, Patrick Campbell, Barry Baker, Mark Cohen, Anton Darmenov, Ravan Ahmadov, Eric James, Edward Hyer, and Peng Xian.
\newblock Multiagency ensemble forecast of wildfire air quality in the {United States}: Toward community consensus of early warning.
\newblock \emph{Bulletin of the American Meteorological Society}, 105\penalty0 (6):\penalty0 E991 -- E1003, 2024.

\bibitem[Liu et~al.(2016)Liu, Mickley, Sulprizio, Dominici, Yue, Ebisu, Anderson, Khan, Bravo, and Bell]{Liu2016particulate}
Jia~Coco Liu, Loretta~J. Mickley, Melissa~P. Sulprizio, Francesca Dominici, Xu~Yue, Keita Ebisu, Georgiana~Brooke Anderson, Rafi F.~A. Khan, Mercedes~A. Bravo, and Michelle~L. Bell.
\newblock Particulate air pollution from wildfires in the western {US} under climate change.
\newblock \emph{Climatic Change}, 138\penalty0 (3):\penalty0 655--666, 2016.

\bibitem[Lynch et~al.(2016)Lynch, Reid, Westphal, Zhang, Hogan, Hyer, Curtis, Hegg, Shi, Campbell, Rubin, Sessions, Turk, and Walker]{gmd-9-1489-2016}
P.~Lynch, J.~S. Reid, D.~L. Westphal, J.~Zhang, T.~F. Hogan, E.~J. Hyer, C.~A. Curtis, D.~A. Hegg, Y.~Shi, J.~R. Campbell, J.~I. Rubin, W.~R. Sessions, F.~J. Turk, and A.~L. Walker.
\newblock An 11-year global gridded aerosol optical thickness reanalysis (v1.0) for atmospheric and climate sciences.
\newblock \emph{Geoscientific Model Development}, 9\penalty0 (4):\penalty0 1489--1522, 2016.

\bibitem[{National Oceanic and Atmospheric Administration}(2016)]{naqfc2016report}
{National Oceanic and Atmospheric Administration}.
\newblock National air quality forecast capability: Updates to operational {CMAQ PM2.5} predictions and ozone predictions, 2016.

\bibitem[Schultz et~al.(2008)Schultz, Heil, Hoelzemann, Spessa, Thonicke, Goldammer, Held, Pereira, and van Het~Bolscher]{schultz2008global}
Martin~G Schultz, Angelika Heil, Judith~J Hoelzemann, Allan Spessa, Kirsten Thonicke, Johann~G Goldammer, Alexander~C Held, Jose~MC Pereira, and Maarten van Het~Bolscher.
\newblock Global wildland fire emissions from 1960 to 2000.
\newblock \emph{Global Biogeochemical Cycles}, 22\penalty0 (2), 2008.

\bibitem[Stajner et~al.(2012)Stajner, Davidson, Byun, McQueen, Draxler, Dickerson, and Meagher]{stajner2012naqfcpm}
Ivanka Stajner, Paula Davidson, Daewon Byun, Jeffery McQueen, Roland Draxler, Phil Dickerson, and James Meagher.
\newblock {US} national air quality forecast capability: Expanding coverage to include particulate matter.
\newblock In Douw~G. Steyn and Silvia Trini~Castelli, editors, \emph{Air Pollution Modeling and its Application XXI}, pages 379--384. Springer Netherlands, 2012.

\bibitem[Tong(2023)]{tong2023haqes}
Daniel Tong.
\newblock {HAQES 3-Hourly Ensemble mean surface total PM2.5 concentration, North America V1.0}.
\newblock Goddard Earth Sciences Data and Information Services Center (GES DISC),, 2023.

\bibitem[{United States Census Bureau}(2023)]{census}
{United States Census Bureau}.
\newblock Gazetteer files: Urban areas, 2023.

\bibitem[{US Environmental Protection Agency}(2024)]{airnow2024}
{US Environmental Protection Agency}.
\newblock {Air Quality System} data mart, 2024.

\bibitem[Yao et~al.(2013)Yao, Brauer, and Henderson]{yao_evaluation_2013}
Jiayun Yao, Michael Brauer, and Sarah~B. Henderson.
\newblock Evaluation of a {Wildfire} {Smoke} {Forecasting} {System} as a {Tool} for {Public} {Health} {Protection}.
\newblock \emph{Environmental Health Perspectives}, 121\penalty0 (10):\penalty0 1142--1147, October 2013.

\bibitem[Ye et~al.(2021)Ye, Arab, Ahmadov, James, Grell, Pierce, Kumar, Makar, Chen, Davignon, et~al.]{ye2021evaluation}
Xinxin Ye, Pargoal Arab, Ravan Ahmadov, Eric James, Georg~A Grell, Bradley Pierce, Aditya Kumar, Paul Makar, Jack Chen, Didier Davignon, et~al.
\newblock Evaluation and intercomparison of wildfire smoke forecasts from multiple modeling systems for the 2019 williams flats fire.
\newblock \emph{Atmospheric Chemistry and Physics Discussions}, 2021:\penalty0 1--69, 2021.

\bibitem[Zhang et~al.(2022)Zhang, Montuoro, McKeen, Baker, Bhattacharjee, Grell, Henderson, Pan, Frost, McQueen, et~al.]{zhang2022development}
Li~Zhang, Raffaele Montuoro, Stuart~A McKeen, Barry Baker, Partha~S Bhattacharjee, Georg~A Grell, Judy Henderson, Li~Pan, Gregory~J Frost, Jeff McQueen, et~al.
\newblock Development and evaluation of the aerosol forecast member in the national center for environment prediction ({NCEP})'s global ensemble forecast system ({GEFS-Aerosols v1}).
\newblock \emph{Geoscientific Model Development}, 15\penalty0 (13):\penalty0 5337--5369, 2022.

\end{thebibliography}
